\documentclass[conference]{IEEEtran}
\IEEEoverridecommandlockouts
\usepackage{cite}
\usepackage{amsmath,amssymb,amsfonts}
\usepackage{graphicx}
\usepackage{textcomp}
\usepackage{xcolor}
\def\BibTeX{{\rm B\kern-.05em{\sc i\kern-.025em b}\kern-.08em
    T\kern-.1667em\lower.7ex\hbox{E}\kern-.125emX}}

\usepackage{algorithm}
\usepackage[noend]{algpseudocode}

\usepackage{graphicx}
\usepackage{xcolor}

\begin{document}

\title{Investigating Semi-Supervised Learning Algorithms in Text Datasets\\
}

\author{\IEEEauthorblockN{1\textsuperscript{st} H. Toprak Kesgin}
\IEEEauthorblockA{\textit{Department of Computer Engineering} \\
\textit{Yildiz Technical University} Istanbul, Turkey \\
tkesgin@yildiz.edu.tr}
\and
\IEEEauthorblockN{2\textsuperscript{nd} M. Fatih Amasyali}
\IEEEauthorblockA{\textit{Department of Computer Engineering} \\
\textit{Yildiz Technical University} Istanbul, Turkey \\
amasyali@yildiz.edu.tr}
}
\maketitle
\begin{abstract}
Using large training datasets enhances the generalization capabilities of neural networks.
Semi-supervised learning (SSL) is useful when there are few labeled data and a lot of unlabeled data.
SSL methods that use data augmentation are most successful for image datasets.
In contrast, texts do not have consistent augmentation methods as images.
Consequently, methods that use augmentation are not as effective in text data as they are in image data.
In this study, we compared SSL algorithms that do not require augmentation; these are self-training, co-training, tri-training, and tri-training with disagreement.
In the experiments, we used 4 different text datasets for different tasks.
We examined the algorithms from a variety of perspectives by asking experiment questions and suggested several improvements.
Among the algorithms, tri-training with disagreement showed the closest performance to the Oracle; however, performance gap shows that new semi-supervised algorithms or improvements in existing methods are needed.
\end{abstract}


\section{Introduction}
Larger training datasets increase the generalizability of neural networks.
If a large amount of data is used, it often outperforms other traditional methods of machine learning.
However, data collection and labeling are laborious and costly.
It may not always be possible to train neural networks with enough samples.
Sometimes, data is readily available, but labeling can be costly.
For example, labeling text or image data is more expensive than acquiring it.

Semi-supervised learning (SSL) is a machine learning method that uses only a few labeled data and a relatively large number of labeled data. As its name suggests, SSL lies between supervised and unsupervised learning.
SSL aims to outperform supervised learning, which uses only labeled samples, and unsupervised learning, in which no samples are labeled.

SSL is gaining popularity with the aim of reducing the amount of labeled data required for neural networks.
It can be achieved by directly using or adapting existing methods for neural networks.
SSL uses supervised learning techniques to generate training data.

There are many methods and techniques for SSL.
The first of these is the consistency regularization method \cite{radford2015unsupervised, springenberg2015unsupervised}.
This method assumes that adding small noise to data points should not have a significant impact on label estimates.
Then it is possible to add unlabeled data points in which the predictions do not change much to a training dataset.

Some SSL methods use generative models \cite{odena2016semi,goodfellow2014generative,kingma2013auto}.
Using such methods, new data points are created from the original data points for training purposes.
Methods like these can be applied to datasets where augmentation can be performed successfully.
Although data augmentation techniques provide significant improvements to image datasets, they do not perform as well with text datasets.

Proxy-label methods are one of the main SSL techniques that don't need data augmentation. \cite{yarowsky1995unsupervised,mcclosky2006effective,mihalcea2004co, nigam2000analyzing,zhou2005tri}
The proxy label method relies on predicting the label of unlabeled data from labeled data.
After that, the predicted samples are combined with the samples with known labels for training.
Therefore, extra information is provided to the model even if the predicted labels are noisy or do not reflect the ground truth.

Proxy label methods differ according to how the labels are predicted.
Self-training is the training of a model with known-label samples, then using this model to iteratively incorporate the label predictions of unknown-label samples into the training.
Self training is one of the earliest methods of SSL.
Another proxy label method, co-training, uses two sets of features that are conditionally independent to predict labels.
In other words, two different models are trained with two different feature sets for the same dataset.
Models provide labels to each other when one's predictions have a high level of confidence, while the other's do not.
As a result, training can continue with the addition of new data points with labels from a different perspective.

Tri-training is another method in proxy label methods.
In tri-training, three different models are trained simultaneously.
If the predictions of the two models are consistent with each other, this label is given to the third model.
Tri-training improves its label prediction accuracy by using multiple models, but it can be computationally expensive since it trains three different models.

The purpose of this study was to compare the performance of proxy label methods that do not require augmentation for text data.
We analyze some situations in the training of algorithms using various questions and answer them in the experiments section.
The performance of the methods was compared according to their supervised and oracle performances on text datasets for four different tasks.
Supervised learning refers to training using only a small number of samples with known labels.
Oracle, on the other hand, refers to the state where all instances have labels.
Oracle represents the upper limit of semi-supervised performance.
It is the aim of SSL to deliver better results than supervised learning, as close to the oracle level as possible.

\section{Literature Review}

Generative SSL methods usually include generative adversarial networks (GANs) \cite{goodfellow2014generative} and variational Auto Encoders (VAEs) \cite{kingma2013auto}.
With GANs, real data points can be used to learn distributions of samples and generate high-quality samples.
Samples produced in this manner can be used directly as labeled for SSL training or can be used to regularize the SSL training model.
As an example, SGAN modifies the GAN architecture in order to also produce a class label using the discriminator network \cite{odena2016semi}.
The proposed SGAN architecture improves classification performance over the non-generative component case.
Additionally, SGAN improves the quality of the generated samples and reduces the generation time.

$\Pi$ Model, which is one of the consistency regularization methods, creates two random augmentation methods for labeled and unlabeled data \cite{sajjadi2016regularization}.
For unlabeled data, randomness in augmentation, drop out, and random max pooling can lead to different predictions.
The consistency of these different estimates establishes the confidence of the $\Pi$ model in unlabeled data.
Other consistency editing methods include Ladder Network\cite{rasmus2015semi,pezeshki2016deconstructing}, Temporal Ensembling \cite{laine2016temporal}, Mean Teacher\cite{tarvainen2017mean}.

One of the oldest SSL methods, self training \cite{yarowsky1995unsupervised, mcclosky2006effective}, has shown success in various text classification tasks like sentiment analysis \cite{he2011self,van2016predicting} and part of speech tagging \cite{van2017normalize}.

Co-training, a method used to train multiple models, has been proven to be effective in a variety of fields\cite{mihalcea2004co, nigam2000analyzing}.
Several studies have found that the training algorithm has an impact on co-training performance.
There is a significant performance improvement with co-training when using support vector machines in the email classification task \cite{kiritchenko2001email}.
Tri-training is similar to co-training; however, trains three classifiers and provides significant improvements in the web page classification task and UCI datasets \cite{zhou2005tri}.
An extension of tri-training for neural networks is Tri-net \cite{dong2018tri}.
Authors found that Tri-net performs significantly on the CIFAR-10 dataset with only a few labeled samples.

In this paper, we used text datasets and classified them using Bert vectors and neural networks.
SSL methods have been explained and discussed from a variety of perspectives in the methods section.
In the experiments section, we discussed SSL with asking and answering some unexamined questions.

\section{Methods}
We explain in this section the SSL algorithms and their extensions used in our experiments. 
Self training is based on combining labeled and unlabeled data.
Unlabeled samples with high confidence scores are included in the training if the estimation is above a certain threshold.
The threshold value is based on the model prediction's confidence value, which is a probability value between 0 and 1.
When this threshold is high, fewer unlabeled samples are included in training.
Also, due to the fact that the model already predicts with high confidence, the impact of these samples on the model's weights is limited.
When this threshold is kept low, incorrectly labeled samples can be included in the training since the label's confidence rate is low, and this will adversely affect the model's performance.

In most cases, this threshold is a single value and samples higher than a certain confidence level are selected.
However, the threshold value does not have to be a single value.
By including samples that are smaller than a certain confidence interval, the model's performance may be improved, since it will provide differentiation in training.
The pseudocode for the threshold-based self-training algorithm is given in Algorithm-1.
In Algorithms 1 and 2, $D$ represents the labeled dataset, $U$ represents the unlabeled dataset, $m$ represents the model to be trained, and $\tau_1$ and $\tau_2$ represent threshold values.
\begin{algorithm}\centering
\caption{Threshold-Based Self-training}
\begin{algorithmic}[1]
\State $m \gets train\_model(D)$
\Repeat
	\State $D_i \gets \varnothing$
	\For {$x \in U$}
    	\If {$\max m(x) > \tau_1$ and $\max m(x) < \tau_2$}
        	\State $D_i \gets \{(x, p(x))\}$
    	\EndIf
    \EndFor
    \State $m \gets train\_model(D \cup D_i)$
\Until {All samples are included in the training OR max iteration count is reached}
\end{algorithmic}
\end{algorithm}
It is also possible to define self-training based on the number of samples included rather than thresholds.
For example, the most reliable x samples can be included in every iteration.
We will call this approach count-based self training.
The pseudocode for count-based self training is found in Algorithm-2.
\begin{algorithm}\centering
\caption{Count-Based Self-training}
\begin{algorithmic}[1]
\State $m \gets train\_model(D)$
\Repeat
	\State $D_i \gets \varnothing$
    \State Sort $U$ from greatest to least according to model's confidence scores
    \State $D_i$ = $U[th1..th2]$
    \State $m \gets train\_model(D \cup D_i)$
\Until {All samples are included in the training OR max iteration count is reached}
\end{algorithmic}
\end{algorithm}
Some algorithms may need to be trained from scratch with new datasets, but neural networks allows the model to be trained where it left off with new-different datasets.

Therefore, the train model function in the 7th line of the algorithms allows you to train a neural network in both directions from the beginning and from where you left off in the previous model. 
In the experiments section, we examined the effects of these two different cases.

Two main differences between co-training and self-training are that co-training requires two different sets of features for training set and uses two models for training.
These two feature sets must be sufficient to train a model and must be conditionally independent.
Co-training involves two feature sets and two models, which allows the models to view the dataset from different perspectives.
Co-training includes samples in training that have a low confidence value for one model and a high confidence value for the other model.
As a result, more information can be added to the model and better models can be trained.
The pseudocode for co-training is found in Algorithm-3.
where $D_1$ and $D_2$ are two different feature sets for the training set, $U$ unlabeled dataset with the same features, $\tau$ is the threshold $m_i$ is the $i^{th}$ model,  $m(x)$ model's confidence scores for each class, and $p_i(x)$ is the label that was predicted by the $i^{th}$ model for the sample $x$.

\begin{algorithm}
\caption{Co-training }
\begin{algorithmic}[1]
\State $m_1 \gets train\_model(D_1)$
\State $m_2 \gets train\_model(D_2)$
\Repeat
    \State $L_1 \gets \varnothing$
    \State $L_2 \gets \varnothing$
	\For {$x \in U$}
    	\If {$\max m_1(x) > \tau \: {\bf and} \: \max m_2(x) < \tau   $}
           	\State $L_1 \gets \{(x, p_1(x))\}$
     \EndIf
     \If {$\max m_2(x) > \tau \: {\bf and} \: \max m_1(x) < \tau   $}
           	\State $L_2 \gets \{(x, p_2(x))\}$
     \EndIf
    \EndFor
\State $m_1 \gets train\_model(D_1 \cup L_2)$
\State $m_2 \gets train\_model(D_2 \cup L_1)$
\Until {no more predictions are confident based on \emph{one} classifier}
\end{algorithmic}
\end{algorithm}

Three models are simultaneously trained in tri-training.
If both models agree on the label of a data point, then the third model includes that data point in its training set.
Tri-training with disagreement adds the condition that the third model makes a different prediction than the two other models.
Unlike co-training, tri-training does not require two sets of features.
Tri-training models differentiate models using bootstrap sampling method and randomness in model training.
When two models agree on a sample's label, it increases confidence over a single model.
After training is complete, the 3 models are combined to provide a final prediction.
The pseudocode for tri-training and tri-training with disagreement is given in Algorithm 4. Tri learning with disagreement is indicated by between square brackets ( $[ ]$ ).
where $D$ is the training set, $U$ is an unlabeled dataset, $m_i$ is the $i^{th}$ model,  $m(x)$ model's confidence scores for each class, and $p_i(x)$ is the label predicted by the $i^{th}$ model for sample $x$.

\begin{algorithm}
\caption{Tri-training [With Disagreement]} 
\begin{algorithmic}[1]
\For {$i \in \{1..3\}$}
\State $S_i \gets bootstrap\_sample(D)$
\State $m_i \gets train\_model(S_i)$
\EndFor
\Repeat
	\For {$i \in \{1..3\}$}
        \State $D_i \gets \emptyset$
		\For {$x \in U$}
            \If {$p_j(x) = p_k(x)(j,k \neq i)$ \par
            \hskip\algorithmicindent [ and $p_i(x) != p_k(x)$ ] }
            	\State $D_i \gets D_i \cup \{(x, p_j(x))\}$
            \EndIf
        \EndFor
       \State $m_i \gets train\_model(D \cup D_i)$
	\EndFor
\Until {none of $m_i$ changes}
\State apply majority vote over $m_i$
\end{algorithmic}
\end{algorithm}

\section{Experiments}

\subsection{Datasets}
Neural network's generalization capacity and performance over a test set increases as the size of the training set increases. 
Figure 1 experimentally illustrates this on the 1150 news\cite{1150news} dataset.
\begin{figure}[h]
\centering
\includegraphics[width=\columnwidth,scale=0.25]{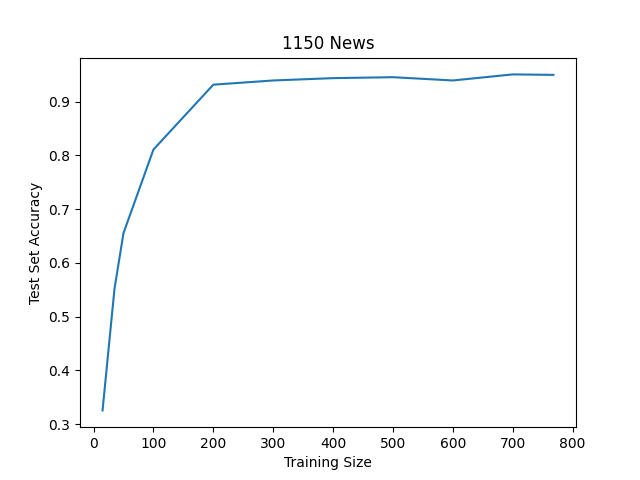}
\caption{Effect of dataset size on accuracy}
\end{figure}

Thus, SSL algorithms have a meaningful effect when there is little training data.
When SSL is applied with enough training data, the increase in accuracy is very low.
This will make it difficult to understand the effects of the algorithms.
For these reasons, we chose datasets with a small number of training samples or included subsets of large sample datasets.
In the experiments, we used text datasets, and we took into account the variety of tasks when selecting the datasets. A summary of the datasets can be found in Table 1.

Turkish movie sentiment dataset (Sent) \cite{sentiment}  contains comments and scores for movies. We reduced this data set by randomly selecting 500 positive 500 negative 1000 comments.
Sarcasm Headlines Dataset (Sarcasm) \cite{sarcasm,book}  also reduced into 1000 sample which 500 sarcastic and 500 non-sarcastic samples. The selected examples were automatically translated into Turkish using Google Translate.
\begin{table}[]
\caption{Datasets summary}
\centering
\begin{tabular}{|l|l|l|l|l|}
\hline
Dataset   & Sample & Class & Task                       \\
\hline
1150 news \cite{1150news} & 1150          & 5           & News classification        \\
Hate \cite{mayda2021turkcce}     & 1000          & 2           & Hate Speech Classification \\
Sent \cite{sentiment}      & 1000          & 2           & Sentiment Analysis         \\
Sarcasm \cite{sarcasm,book}   & 1000          & 2           & Sarcasm detection         \\
\hline
\end{tabular}
\end{table}
In the experiments, each text was represented with 768 length vectors using the Berturk model \cite{stefan_schweter_2020_3770924} and training was performed with these vectors.
\subsection{Training Settings}
In all experiments, the same neural network was used as a classifier.
In the first layer of the neural network, there are 768 neurons, which are the input dimensions.
Two hidden layers, each consist of 16 neurons, are then connected to this layer.
Relu is the activation function of these layers.
The last layer is determined by the number of classes in the dataset.
Softmax is the activation function of the last layer that gives confidence scores for each class of predictions.
Adam \cite{kingma2014adam} algorithm is used while optimizing the models.
Neural networks are stochastic algorithms.
Therefore, they may produce different results on different runs of the same dataset.
Different weights and randomness in the optimizer affect the classifier's performance.
For these reasons, three-fold cross-validation was used when evaluating the models.
After each fold was created, some part of the unlabeled data set was selected based on the separation rate.
The models for each fold were trained five times with different weights.
The maximum test accuracy achieved by the models throughout the training was accepted as the success of the relevant training.
For each evaluation, 3-fold times 5 trials 15 trainings were performed for each model.
We used the t-test with $p = 0.10$ to test the significance of these 15 results between models.
In the tables, Abbreviations as follows; Threshold Based Self Training (TBST), Count Based Self Training (CBST), Tri-training (TT), Tri-training with Disagreement (TTWD) and Co-Training (CT).
In all tables, green indicates a statistically significant improvement according to the supervised algorithm, where only labeled data were used.
A red color indicates a statistically significant decline.

\subsection{How well semi-supervised algorithms work?}

In this section, we compare the models without making any hyperparameter adjustments.
In threshold-based self-training, we included samples with confidence scores greater than 0.90 from unlabeled samples in the training set at each iteration.
For count-based self-training, we included 100 of the most reliable unlabeled samples in the training set at each iteration.
The same algorithm 4 is used for Tri-training and Tri-training with disagreement.
For co-training, the feature set was divided into two equal parts.
In other words, the first model used the first 384 features, and the second model used the last 384 features.
Ensemble models were used for the evaluation of tri-training, tri-training with disagreement, and co-training.
In each iteration, the models were optimized by continuing from where they left off in the previous iteration.
Table 2 shows performance comparisons between algorithms.
Oracle specifies the state where all samples in the training dataset have been labelled, and the maximum performance limit for algorithms.
Supervised, on the other hand, refers to a situation in which only labeled examples are used, and unlabeled examples are not included in the training.
\begin{table}[h]
\caption{Comparison of Algorithms Baselines}
\centering
\begin{tabular}{|lllll|}
\hline
\multicolumn{5}{|c|}{Unlabeled sample rate $0.95$} \\
\hline
           & News                         & Sent                    & Sarcasm & Hate \\ \hline
Oracle     & 95.08                        & 87.36                        & 76.88   & 78.96       \\
Supervised & 86.99                        & 81.52                        & 65.49   & 72.70       \\
TBST       & {\color[HTML]{FE0000} 83.74}                        & 81.73                        & 64.79   & 71.49       \\
CBST       & 85.97                        & 80.69                        & 65.43   & 71.83       \\
TT         & 87.67 & {\color[HTML]{009901} 83.05} & 65.71   & 71.81       \\
TTWD       & {\color[HTML]{009901} 90.78} & 82.92 & 66.32   & 73.55       \\
CT         & 88.37 & 82.61 & 65.04   & 72.79  
\\
\hline
\multicolumn{5}{|c|}{Unlabeled sample rate $0.90$} \\
\hline
           & News                         & Sent                   & Sarcasm & Hate \\ \hline
Oracle     & 95.08                        & 87.36                        & 76.88   & 78.96       \\
Supervised & 90.54                        & 83.29                        & 68.53   & 73.98       \\
TBST       & 90.54                        & 84.01                        & 68.32   & 72.94       \\
CBST       & 90.14                        & 82.88                        & 68.04   & 73.68       \\
TT         & {\color[HTML]{009901} 92.99} & {\color[HTML]{009901} 84.72} & 68.13   & 73.09       \\
TTWD       & {\color[HTML]{009901} 93.67} & {\color[HTML]{009901} 84.49} & 69.44   & 74.89       \\
CT         & {\color[HTML]{009901} 92.33} & {\color[HTML]{009901} 84.39} & 68.11   & 74.53  
\\
\hline 
\multicolumn{5}{|c|}{Unlabeled sample rate $0.80$} \\
\hline
           & News                         & Sent                    & Sarcasm & Hate \\ \hline
Oracle     & 95.08                        & 87.36                        & 76.88   & 78.96       \\
Supervised & 93.64                        & 85.19                        & 71.12   & 76.21       \\
TBST       & 93.77                        & 85.51                        & 70.83   & 75.98       \\
CBST       & 93.11                        & 84.63                        & 70.56   & 75.60       \\
TT         & 94.19 & 85.63 & 70.97   & 76.38       \\
TTWD       & {\color[HTML]{009901} 94.54} & 85.47 & 71.12   & 76.36       \\
CT         & 93.76 & 85.56 & 71.32   & 76.62  
\\
\hline

\end{tabular}
\\
\end{table}
We conducted a t-test between the supervised and semi-supervised algorithms.
Green indicates significant improvements in the supervised algorithm, while red indicates worsening.
Among the algorithms, tri-training with disagreement is the most promising.
When the unlabeled dataset is 90\%, significant improvements are achieved in the two datasets compared to supervised learning.
On the other hand, self training algorithms cannot surpass supervised learning algorithm.
\subsection{How do sampling methods affect tri-training?}
Tri-training trains 3 models with labeled dataset before iterative learning begins.
It is important for different sampling strategies to differentiate models in tri-training, which is based on the differentiation of model predictions.
The bootstrap method is used to differentiate the predictions of these models.
In this study, we compared five different sampling strategies for TT and TTWD.
As part of choosing these strategies, we decided how many samples will be selected based on the number of labeled samples and whether we will select with or without replacement.
We will call $x$ the number of samples in the labeled dataset.
\begin{table}[h]
\caption{Comparison of sampling strategies for tri-training}
\centering
\begin{tabular}{|llllll|}
\hline
\multicolumn{6}{|c|}{TT - Unlabeled Sample Rate 0.90} \\
\hline
Size & Replacement                    & News                          & Sent                     & Sarcasm                       & Hate                   \\ \hline
x           & False                          & {\color[HTML]{70AD47} 92.99} & {\color[HTML]{70AD47} 84.72} & 68.13                        & 73.09                        \\
2x          & True                           & {\color[HTML]{70AD47} 93.39} & {\color[HTML]{70AD47} 84.40} & 68.41                        & 74.43                        \\
x           & True                           & {\color[HTML]{70AD47} 92.80} & {\color[HTML]{70AD47} 84.45} & 68.29                        & 73.00                        \\
x/2         & True                           & 88.71                        & 84.31                        & {\color[HTML]{FE0000} 66.52} & {\color[HTML]{FE0000} 70.96} \\
x/3         & No intersection & {\color[HTML]{FE0000} 83.22} & {\color[HTML]{70AD47} 84.76} & 67.40                        & {\color[HTML]{FE0000} 70.85}
\\
\hline
\multicolumn{6}{|c|}{TTWD - Unlabeled Sample Rate 0.90} \\
\hline
Size & Replacement                    & News                          & Sent                    & Sarcasm & Hate \\
x           & False                          & {\color[HTML]{70AD47} 93.67} & {\color[HTML]{70AD47} 84.49} & 69.44  & 74.89      \\
2x          & True                           & {\color[HTML]{70AD47} 93.95} & {\color[HTML]{70AD47} 84.71} & 69.61  & 74.77      \\
x           & True                           & {\color[HTML]{70AD47} 93.90} & {\color[HTML]{70AD47} 84.80} & 69.01  & 74.64      \\
x/2         & True                           & {\color[HTML]{70AD47} 93.31} & {\color[HTML]{70AD47} 84.84} & 68.85  & 74.51      \\
x/3         & No intersection & {\color[HTML]{70AD47} 93.24} & {\color[HTML]{70AD47} 85.00} & 68.49  & 74.64  
\\
\hline
\end{tabular}
\\
\end{table}
Table 3 presents a comparison of tri-training algorithms for different sampling strategies when initially training the models.
Size is the number of samples selected; replace is whether replacement is performed with or without sampling.
On the other hand, no intersection means that the data set is divided into three subsets that do not intersect.
The sampling strategies were used to train the models before the iterative process started.
It is a replacement for the second line of Algorithms 3 and 4.
When the results are examined, it is clear that although the decrease in the number of samples selected for TT will differentiate the models, it will adversely affect the initial performance of the models.
On the contrary, TTWD does not degrade performance when the number of selected samples is decreased.
The disagreement condition in this TTWD improves the poor initial performance of the models.
It does not significantly affect the results of either algorithm if the sampling is done with or without replacement.

\subsection{Is it better to train a model from scratch every time?}
The weights of neural networks are randomly initialized at the beginning of training.
Using SSL algorithms, the trained model is used to predict labels for unlabeled samples, and some of these samples are added to the training set based on certain criteria.
After unlabeled samples are included in the training set, the model must be (re)trained.
In the previous sections, we continued training using the same neural network model.
However, with the new samples, it is possible to train a new model with different initial weights.
The purpose of this section is to examine the effect of randomly initializing the weights of neural networks during SSL training.
In other words, it is a replacement for the \emph{train\_model} function in the algorithms in the methods section.
\begin{table}[h]
\caption{Effect of creating a new model at each iteration}
\centering
\begin{tabular}{|llllll|}
\hline
\multicolumn{6}{|c|}{Unlabeled Sample Rate 0.90} \\
\hline
Alg  & New Model & News                          & Sent                          & Sarcasm & Hate   \\
TBST & True      & {\color[HTML]{70AD47} 93.98} & {\color[HTML]{70AD47} 85.77} & 69.84  & 74.49 \\
TBST & False     & 90.54                        & 84.01                        & 68.32  & 72.94 \\
CBST & True      & {\color[HTML]{70AD47} 92.80} & {\color[HTML]{70AD47} 85.19} & 70.24  & 75.19 \\
CBST & False     & 90.14                        & 82.88                        & 68.04  & 73.68 \\
TT   & True      & {\color[HTML]{70AD47} 93.67} & {\color[HTML]{70AD47} 85.31} & 68.97  & 74.38 \\
TT   & False     & {\color[HTML]{70AD47} 92.99} & {\color[HTML]{70AD47} 84.72} & 68.13  & 73.09 \\
TTWD & True      & {\color[HTML]{70AD47} 92.40} & {\color[HTML]{70AD47} 84.60} & 68.77  & 74.70 \\
TTWD & False     & {\color[HTML]{70AD47} 93.67} & {\color[HTML]{70AD47} 84.49} & 69.44  & 74.89 \\
CT  & True      & {\color[HTML]{70AD47} 93.39} & {\color[HTML]{70AD47} 85.37} & 69.19  & 74.85 \\
CT  & False     & {\color[HTML]{70AD47} 92.33} & {\color[HTML]{70AD47} 84.39} & 68.11  & 74.53
\\
\hline
\end{tabular}
\\
\end{table}
The performance of self-training improved significantly by training a new model in each iteration.
Two datasets showed statistically significant improvements, while there were no significant improvements before.
In other algorithms, significant improvements in performance were not seen when compared to the new model was not trained.
Even though training models from scratch improves performance in some algorithms, it is very costly in terms of training time.

\subsection{Is it better to use an ensemble model or the best single model?}
The performance of the tri-training algorithms and co-training are evaluated by combining the predictions of the models.
For these multi-model algorithms, we investigated whether the most successful single model or the ensemble of 3 models was more successful.
\begin{table}[h]
\caption{Comparison of ensemble and single most successful model}
\centering
\begin{tabular}{|llllll|}
\hline
\multicolumn{6}{|c|}{Unlabeled Sample Rate 0.90} \\
\hline
Alg  & Eval     & News                          & Sent                          & Sarcasm & Hate   \\
TT   & Ensemble & {\color[HTML]{70AD47} 92.99} & {\color[HTML]{70AD47} 84.72} & 68.13  & 73.09 \\
TT   & Single   & {\color[HTML]{70AD47} 93.34} & {\color[HTML]{70AD47} 85.07} & 68.72  & 73.90 \\
TTWD & Ensemble & {\color[HTML]{70AD47} 93.67} & {\color[HTML]{70AD47} 84.49} & 69.44  & 74.89 \\
TTWD & Single   & {\color[HTML]{70AD47} 93.41} & {\color[HTML]{70AD47} 84.60} & 69.65  & 75.02 \\
CT & Ensemble & {\color[HTML]{70AD47} 92.33} & {\color[HTML]{70AD47} 84.39} & 68.11 & 74.53 \\
CT & Single & 90.75 & 83.68 & 67.41 & 73.87
\\
\hline
\end{tabular}
\\
\end{table}
The single most successful model in tri-training showed slight improvements over the ensemble model.
In tri-training with disagreement, the performance changes based on the dataset.
In contrast, the ensemble model clearly performed better in co-training.

\subsection{How different threshold values affect self training?}
In self training the model is trained using labeled data first.
This model makes predictions on the unlabeled dataset, and those whose confidence scores exceed a certain threshold value are trained together with the labeled ones.
The threshold value affects the performance of self-training.
Although a high threshold increases the reliability of the labels, it will not change the current weights of the models much.
By contrast, choosing a low threshold reduces label reliability.
By limiting the threshold from the top, the model will be able to change its current state with greater margins.
\begin{table}[h]
\caption{Comparison of sampling strategies for tri-training}
\centering
\begin{tabular}{|llllll|}
\hline
\multicolumn{6}{|c|}{Threshold Based Self Training - Unlabeled Sample Rate 0.90} \\
\hline
Th1 & Th2 & News                          & Sent   & Sarcasm & Hate                          \\ \hline
0.7 & 1.0 & 90.45                        & 84.08 & 68.08  & 72.60                        \\
0.8 & 1.0 & 90.57                        & 83.95 & 68.27  & {\color[HTML]{FF0000} 72.15} \\
0.9 & 1.0 & 90.54                        & 84.01 & 68.32  & 72.94                        \\
0.7 & 0.9 & {\color[HTML]{70AD47} 91.86} & 83.64 & 68.65  & 73.41                        \\
0.7 & 0.8 & {\color[HTML]{70AD47} 92.23} & 83.53 & 68.69  & 73.70                        \\
0.8 & 0.9 & {\color[HTML]{70AD47} 91.84} & 83.69 & 68.67  & 73.45                       
\\
\hline
\multicolumn{6}{|c|}{Count Based Self Training - Unlabeled Sample Rate 0.90} \\
\hline
Th1 & Th2 & News   & Sent   & Sarcasm & Hate   \\ \hline
0 & 300 & 90.64 & 82.84 & 68.03  & 73.94 \\
0 & 200 & 90.68 & 82.92 & 68.12  & 73.70 \\
0 & 100 & 90.14 & 82.88 & 68.04  & 73.68 \\
100 & 200 & 90.73 & 82.99 & 68.16  & 73.77 \\
100 & 300 & 90.66 & 83.12 & 68.17  & 73.87 \\
200 & 300 & 90.59 & 83.16 & 68.17  & 73.74
\\
\hline
\end{tabular}
\\
\end{table}
In Table 6, we compare two different self-training algorithms and different threshold values for self-training.
Generally, threshold-based self training outperforms count-based self training.
Additionally, instead of only including samples below a certain confidence level in the table, not including samples below a certain confidence level provides various improvements, as it creates differences in model training.

\section*{Conclusions}
We compared SSL algorithms that do not require data augmentation across a variety of text datasets.
We discuss the algorithms from various perspectives.
In the experiments section, we examined the effects of trade-offs of algorithms with different questions.
We have proposed several improvements to some algorithms.
Our findings are summarized as follows.
\begin{itemize}
  \item Although SSL methods have been able to increase success in some datasets, they have not been effective in all cases.
  \item Tri-training with disagreement is the most promising among SSL algorithms.
  \item A small sample size reduces the success rate compared with a large sample size in initial training of tri-training.
  \item Despite its higher cost, the success rate of self-training increases significantly when a new model is trained from scratch for each iteration.
  \item Instead of one, the use of two thresholds that limit the confidence value at the top and bottom increases the effectiveness of self-training.
  \item In tri-training, the best single model and the best ensemble of three models are no difference in success.
\end{itemize}
Tri-training with disagreement is the most promising, and algorithms that use more than one model are generally more effective.
However, it seems that due to a difference between existing methods and Oracle's performance, in which the label of all unlabeled samples is correctly known, new SSL algorithms or improvements in existing methods are waiting to be found.

\section*{Acknowledgment}

This study was supported by the Scientific and Technological Research Council of Turkey (TUBITAK) Grant No: 120E100.

\bibliography{citation}
\bibliographystyle{IEEEtran}

\end{document}